%% file: main.tex
\documentclass[fleqn,10pt]{wlscirep}
\usepackage[utf8]{inputenc}
\usepackage[T1]{fontenc}
\usepackage{lineno}
\usepackage{amsmath,amssymb,amsfonts}
\usepackage{graphicx}
\usepackage{caption}
\usepackage{subcaption}
\usepackage{multirow}
\usepackage{float}
\usepackage{physics}
\usepackage{tikz}
\usepackage{spverbatim}
\usepackage{listings}
\usepackage{xcolor}
\usepackage{framed}
\usetikzlibrary{decorations.pathreplacing}
\usetikzlibrary{patterns}
\usetikzlibrary {arrows.meta} 
\usepackage[noend]{algpseudocode}
\usepackage{algorithm2e}
\usepackage{hyperref}

\hypersetup{
    breaklinks=true,    
    colorlinks=true,    
    linkcolor=blue,     
    urlcolor=blue       
}

\lstdefinelanguage{json}{
    basicstyle=\ttfamily\small,
    numbers=left,
    numberstyle=\scriptsize,
    stepnumber=1,
    numbersep=8pt,
    showstringspaces=false,
    breaklines=true,
    frame=lines,
    literate=
     *{0}{{{\color{blue}0}}}{1}
      {1}{{{\color{blue}1}}}{1}
      {2}{{{\color{blue}2}}}{1}
      {3}{{{\color{blue}3}}}{1}
      {4}{{{\color{blue}4}}}{1}
      {5}{{{\color{blue}5}}}{1}
      {6}{{{\color{blue}6}}}{1}
      {7}{{{\color{blue}7}}}{1}
      {8}{{{\color{blue}8}}}{1}
      {9}{{{\color{blue}9}}}{1}
      {:}{{{\color{red}:}}}{1}
      {,}{{{\color{red},}}}{1}
      {\{}{{{\color{brown}\{}}}{1}
      {\}}{{{\color{brown}\}}}}{1}
      {[}{{{\color{brown}[}}}{1}
      {]}{{{\color{brown}]}}}{1},
}

\nolinenumbers

\title{Bitcoin Research with a Transaction Graph Dataset
}

\author[1,2,*]{Hugo Schnoering}
\author[1]{Michalis Vazirgiannis}
\affil[1]{Ecole Polytechnique, Palaiseau, 91120, France}
\affil[2]{Coinshares, Paris, 75002, France}

\affil[*]{corresponding author(s): Hugo Schnoering (hschnoering@coinshares.com)}

\begin{abstract}
Bitcoin, launched in 2008 by Satoshi Nakamoto, established a new digital economy where value can be stored and transferred in a fully decentralized manner - alleviating the need for a central authority. This paper introduces a large scale dataset in the form of a transactions graph
representing transactions between Bitcoin users along with a set of tasks and baselines. The graph includes 252 million nodes and 785 million edges, covering a time span of nearly 13 years of and 670 million transactions. Each node and edge is timestamped. 
As for supervised tasks we provide two labeled sets i.  a 33,000 nodes based on entity type and ii. nearly 100,000 Bitcoin addresses labeled with an entity name and an entity type. This is the largest publicly available data set of bitcoin transactions designed to facilitate advanced research and exploration in this domain, overcoming the limitations of existing datasets. Various graph neural network models are trained to predict node labels, establishing a baseline for future research. In addition, several use cases are presented to demonstrate the dataset's applicability beyond Bitcoin analysis. Finally, all data and source code 
is made publicly available to enable reproducibility of the results.

\end{abstract}

\begin{document}

\flushbottom
\maketitle

\thispagestyle{empty}

\section*{Background \& Summary}

Bitcoin\cite{nakamoto2008bitcoin} is an ensemble of concepts introduced in 2008 by the pseudonymous entity Satoshi Nakamoto, allowing the creation of a groundbreaking new digital economy. Analogous to traditional economies, value within this economy can be stored and transferred between participants. The term 'Bitcoin'  refers to the protocol governing the economy's rules, the peer-to-peer network of participants, the blockchain (a public ledger where all transactions are recorded), or, with a lower case, the unit of value within the economy. Bitcoin differentiates itself from traditional economies by not relying on a central authority to control inflation or validate transactions. Instead, Bitcoin relies on a distributed, peer-to-peer system that verifies adherence to the rules of the economy. Since its inception, Bitcoin has experienced increasing adoption. According to Glassnode\cite{glassnode}, the average daily number of users has reached 270 thousand by 2023. In the same year, approximately 8.6 trillion dollars were transferred using the Bitcoin network. Furthermore, Bitcoin has garnered growing interest from the scientific community, with more than 30 thousand relevant research papers indexed annually on Google Scholar over the past five years\cite{googlescholar}. \\

Despite the public availability of all Bitcoin transaction data, there is a notable scarcity of public datasets available for researchers. The Bitcoin community focuses on improving security, scalability, and utility, while also addressing risks such as security vulnerabilities and financial crimes. Analyzing Bitcoin transaction graphs has become increasingly important and holds significant potential, as these visualizations provide essential insights into the health and growth of the network. Furthermore, studying transaction patterns helps in identifying potential risks, including criminal activities such as money laundering, fraud, and other malicious behaviors. By detecting anomalies within transaction flows, these graphs contribute to a better understanding of the Bitcoin econonomy and assist in mitigating threats. \\

Most existing datasets provide labeled sets of addresses\cite{conti2018economic,paquet2019ransomware}, placing the onus on researchers to construct the graph and associate addresses with graph nodes. While this methodology offers a degree of flexibility, it necessitates that researchers download the blockchain data, extract pertinent information, develop a graph construction methodology, and apply curation and other processing techniques. These processes require highly specialized and advanced knowledge of Bitcoin, representing a significant barrier to entry for researchers aiming to study Bitcoin. Some open-source graph-form datasets available, such as Elliptic 1 and Elliptic 2\cite{weber2019anti,bellei2024shape}, developed and provided by the blockchain forensic firm Elliptic. Elliptic 1 comprises a graph with 204 thousand nodes representing transactions and 234 thousand directed edges depicting the flow of funds between these transactions. Elliptic 2 includes 122 thousand graphs, encompassing a total of 49 million nodes and 196 million edges. Here, the nodes signify users, and the edges denote transactions between these users. Both datasets utilize a binary labeling for the nodes: licit or illicit. While these datasets are particularly conducive to research on money laundering, their applicability to other research areas is limited. \\

In this work, we introduce a very large scale transaction data in the form of a graph. The nodes represent identifiable and tangible organizations, individuals, or institutions that are involved in the Bitcoin network, hereafter referred to as \textit{real entities}. The edges represent value transfers between these entities. This is the largest publicly available data set of bitcoin transactions designed to facilitate advanced Bitcoin research and exploration, overcoming the limitations of existing datasets.
It includes 252 million nodes and 785 million edges and is based on nearly 13 years of data, representing 670 million transactions. The graph is temporal, with each node and edge timestamped indicating their creation time, providing thus temporal information. The edges are directed representing the direction of value transfers. Additionally, the graph is labeled, with 33K nodes labeled based on the type of the real entity they represent. Our dataset also includes a list of nearly 100,000 Bitcoin addresses labeled with entity names and types, which allows us to label 33,000 nodes in the graph. \\

Additionally, it contains approximately 14 million text posts extracted from the forum Bitcointalk (\url{https://bitcointalk.org/}), which were used to create the list of addresses. Finally, we trained several graph neural networks and a gradient boosting classifier on supervised classification tasks. The results presented in this work establish useful baselines for future research comparisons.\\

The remainder of this work is organized as follows. First, we introduce the comprehensive methodology employed for the construction of the graph, including various curation methods. Next, we detail the node labeling strategy. After constructing the graph, we present the training framework used to train various graph neural network models for predicting node labels. Finally, we present several use cases for this dataset, highlighting its applications beyond Bitcoin research.

\subsection*{Framework of Bitcoin} Bitcoin relies on asymmetric cryptography, with each user possessing several private keys to secure and spend their funds. These private keys are never revealed to the network; instead, pseudonyms are computed from these private keys to identify the user on the network. In the following, the different types of pseudonyms will be collectively referred to as \textit{addresses}. The network’s units of value, bitcoins, are stored in data chunks called transaction outputs (TXOs). A TXO is defined by a value $v$ (in satoshis, 1 satoshi = $10^{-8}$ bitcoin) and a locking script $p$ that specifies the conditions to spend $v$. Generally, $p$ specifies addresses derived from private keys that can spend $v$.  A Bitcoin transaction is characterized by an initial set of TXOs, which are consumed and destroyed to fund the transaction, and a second set, which is created as a result of this transaction. A transaction is valid if the input value is greater than or equal to the output value, if the input UTXOs exist (i.e., they were created by previous transactions) and have not yet been spent. A TXO that has not yet been spent is called an \textit{unspent transaction output} (UTXO). The output TXOs can then be used as inputs for future transactions. A single transaction can gather multiple entities as inputs and outputs. A transaction can thus be viewed as a transformation of one set of TXOs into another, altering the value distribution among entities. A transaction is illustrated in Figure \ref{fig:scheme_transaction}.

\input{transaction_scheme}

\section*{Methods}

\subsection*{Graph Construction}

\subsubsection*{Raw data extraction} All transactions since the inception of the Bitcoin economy are stored in a public ledger called the Bitcoin blockchain. The blockchain is maintained through a decentralized network of peers. Every ten minutes, a block of new transactions is appended. After installing the latest Bitcoin Core software and configuring the node, we set up a Bitcoin Core node. The node allowed us to connect to the network of peers and download the complete transaction ledger. The entire transaction history was saved in the local blockchain data directory, specifically in the 'blkXXXXX.dat' files located within the 'blocks' folder created by the node. Subsequently, we used parsing techniques to extract all transaction details. This process ensured accurate data for our analysis. In this work, we considered the transactions contained within the first 700,000 blocks of the blockchain.

\subsubsection*{Definition of nodes} All circulating bitcoins are allocated within unspent TXOs, each safeguarded by a locking script. Several TXOs can be locked by the same script, and thus be spent by the same address or group of addresses. Furthermore, a transaction can be conceptualized as a transfer of value from one set of scripts to another. In this sense, scripts can be regarded as the owners of the bitcoins they lock. Consequently, locking scripts naturally emerge as candidates for the nodes within the graph dataset. All TXOs and scripts that have existed can be inferred from transaction data. In our analysis, TXOs with a zero value have been excluded, and we have identified over 874 million scripts. \\

A script is derived from a set of private keys held by one or more entities thus making these entities the de facto owners of the bitcoins protected by the script. Typically, a user may possess multiple private keys for purposes of management, security, or privacy. Additionally, the derivation of a locking script from a set of private keys is not unique. As a result, a user generally owns or has owned bitcoins within TXOs protected by different scripts. For a more effective study of Bitcoin, it is preferable to analyze value transfers between real entities or users rather than scripts. This approach has been adopted in most previous research papers preceding this work. Therefore, it is necessary to identify and cluster scripts that likely belong to a single entity, which will then represent a node in the graph.\\

To achieve this, we employed heuristics developed in previous research\cite{schnoering2024assessing}. These heuristics leverage established behavioral patterns and habits of Bitcoin users, along with known human biases, to establish links between scripts appearing in the same transaction. Consequently, the nodes in our graph represent clusters of scripts, with approximately 252 million scripts identified. Each cluster is identified by a unique integer alias. Henceforth, a TXO will be characterized by a value $v$ and the alias $a$ representing the cluster of its locking script.

\subsubsection*{Edges} A transaction is the transformation of a set of input TXOs $\Delta_\text{in}$ into a new set of output TXOs $\Delta_\text{out}$. There is nothing to prevent an alias from being present in both the inputs and the outputs. This situation is common, for example, when receiving change from a payment, as all input TXOs will be fully consumed regardless of the payment amount. Since an alias can appear in both the input and the output of a transaction, it needs to be determined whether this alias sends or receives value during the transaction. We define the value received by an alias $a$ using the equation (\ref{eq:value_received}). This value is simply the difference between the value received in the output and the value spent in the input.

\begin{equation}
    v_\Delta(a) = \sum_{(v^\prime, a^\prime) \in \Delta_\text{out}, \ a^\prime = a} v^\prime - \sum_{(v^\prime, a^\prime) \in \Delta_\text{in}, \ a^\prime = a} v^\prime
    \label{eq:value_received}
\end{equation}

Consequently, the entity denoted as $a$ can be classified as a recipient if the net value received is positive, else $a$ is a sender. The quantity transmitted from sender $a$ to recipient $a^\prime$ is defined as the proportion of the total input value provided by $a$ times the value received by $a^\prime$. An edge is finally drawn between from each sender to each recipient of the considered transactions. 

\subsubsection*{CoinJoin and colored coin transactions} CoinJoin refers to a specific type of transaction that adds a layer of privacy to Bitcoin. The transaction ledger is public, and each transaction can be analyzed, making it relatively simple to follow value flows. Specifically, for a given user, it is straightforward to trace their wealth, from whom they receive value, and to whom they send it, which seriously compromises their privacy. CoinJoin works by combining multiple individual transactions from different users into a single large transaction. Each participant contributes input TXOs and specifies output TXOs without revealing which inputs belong to which outputs. As a result, it obscures the origin and destination of transactions, making it harder for external observers to trace the transfer flow of a particular user. This type of transaction also helps to thwart certain clustering heuristics by leading them to cluster scripts that do not belong to the same user. For all these reasons, we have decided to exclude these transactions in (1) the construction of script clusters and (2) the addition of edges in the graphs. The construction of this type of transaction is made possible through dedicated software, and there are several different implementations available. We will use heuristics developed in previous work\cite{schnoering2023heuristics} to detect these transactions. These heuristics were developed by analyzing the open-source implementations of these softwares to identify recognizable patterns in such transactions. \\

Colored coin transactions are utilized to transfer value in forms other than bitcoin, including other cryptocurrencies and tangible assets. These transactions embed additional information, such as the type of asset or the quantity being transferred, within the transaction's locking script, making them relatively straightforward to identify. To accurately detect these transactions, we devised heuristics based on established protocols like Open Asset, Omni Layer, and EPOBC\cite{colored_coins}. Consequently, we exclude these transactions from our graph construction to maintain the integrity of our analysis by focusing on standard Bitcoin transactions.

\subsubsection*{Attributes} Attributes attached to edges represent the aggregate characteristics of the directed value transfers. Attributes attached to nodes are primarily derived from the edges involving those nodes, providing insights into the nodes' transactional behavior. The different attributes are described in Tables \ref{tab:table_edge} and \ref{tab:table_node}. The blockchain is a chain of blocks, each containing a sequence of transactions. When defining the attributes, the block index of a transaction refers to the index of the block containing the transaction within the blockchain. Consequently, the block index can be regarded as a timestamp.

\begin{table}[H]
\footnotesize
    \centering
    \begin{tabular}{|l|l|l|}
    \hline
    \textbf{Column name}  & \textbf{Description}  \\
    \hline
    a & Node alias of the sender \\ \hline
    b & Node alias of the sender \\ \hline
    reveal & Block index of the first transaction  \\ \hline
    last\_seen & Block index of the last transaction  \\ \hline
    total & Total number of transactions \\ \hline
    min\_sent & Minimum sent in a single transaction \\ \hline
    max\_sent & Maximum sent in a single transaction \\ \hline
    total\_sent & Total sent in all transactions \\ \hline
    \end{tabular}
    \caption{Attributes of the edges.}
    \label{tab:table_edge}
\end{table}

\begin{table}[H]
\footnotesize
\centering
\begin{tabular}{|l|l|l|}
\hline
\textbf{Category} & \textbf{Column name} & \textbf{Description} \\ 
\hline

\multirow{2}{*}{} 
& alias & Identifier of the cluster \\ \cline{2-3} 
& label & Label of the cluster (see 'Node labels') \\ \cline{2-3} \hline

\multirow{3}{*}{Degree} 
& degree & Degree of the node \\ \cline{2-3} 
& degree\_in & The number of incoming edges to the node \\ \cline{2-3} 
& degree\_out & The number of outgoing edges from the node \\ \hline

\multirow{6}{*}{Transfer} 
& total\_transaction\_in & The total count of value transfers received by the node \\ \cline{2-3} 
& total\_transaction\_out & The total count of value transfers initiated by the node \\ \cline{2-3} 
& first\_transaction\_in & Block index of the first transfer received \\ \cline{2-3} 
& last\_transaction\_in & Block index of the last transfer received \\ \cline{2-3} 
& first\_transaction\_out & Block index of the first transfer sent \\ \cline{2-3} 
& last\_transaction\_out & Block index of the last transfer sent \\ \hline

\multirow{6}{*}{Value} 
& min\_sent & The smallest value transferred out in a single transaction \\ \cline{2-3} 
& max\_sent & The largest value transferred out in a single transaction \\ \cline{2-3} 
& total\_sent & The cumulative value of all outgoing transfers \\ \cline{2-3} 
& min\_received & The smallest value received in a single transaction \\ \cline{2-3} 
& max\_received & The largest value received in a single transaction \\ \cline{2-3} 
& total\_received & The cumulative value of all incoming transfers \\ \hline

\multirow{4}{*}{Cluster} 
& cluster\_size & Number of scripts represented by the cluster \\ \cline{2-3} 
& cluster\_num\_edges & Number of value transfers within the cluster  \\ \cline{2-3} 
& cluster\_num\_cc & Number of connected components within the cluster  \\ \cline{2-3} 
& cluster\_num\_nodes\_in\_cc & Number of non-isolated scripts in the cluster \\ \hline

\end{tabular}
\caption{Attributes of the nodes.}
\label{tab:table_node}
\end{table}

\subsection*{Node labels}

 Various real-world entities with distinct motivations utilize Bitcoin, including individuals, government organizations, corporations, service providers, and criminal organizations. Extensive research efforts in Bitcoin dedicate to examining the behavior and dynamics of value transfers among these diverse entities. These studies are invaluable for providing insights into the purposes and motivations behind Bitcoin usage. Bitcoin users are identified by randomly generated addresses. Information from the blockchain alone is insufficient to ascertain the true identity or nature of the entity represented by an alias. In this section, we outline the methodology we used to label addresses and aliases. Our approach predominantly relied on off-chain data, distinguishing it from on-chain data that originates directly from the blockchain. A significant portion of our off-chain data was sourced from BitcoinTalk (\url{https://bitcointalk.org/}), the principal forum for Bitcoin. We employed ChatGPT\cite{brown2020language} to extract Bitcoin addresses and  infer the identity or type of entity associated with each address by analyzing the textual context of the forum messages where the addresses were mentioned. \\

 From the entities previously examined in prior research studies, we have decided to investigate the following types of entities:
 {\small
\begin{itemize}
\item Individual
\item Mining: individual or entity that validates and confirms transactions on the Bitcoin network. 
\item Exchange: online platform that facilitates the buying, selling, and trading of cryptocurrencies.
\item Marketplace: online platform where users can buy and sell goods or services using bitcoin as the primary form of payment.
\item Gambling: online platform where users can wager and play casino games, sports bet, and participate in lotteries using Bitcoin.
\item Bet: address created by a gambling service specifically for receiving funds related to a unique bet.
\item Faucet: promotional tool that rewards users with small amounts of bitcoin for completing tasks or viewing advertisements. 
\item Mixer: service that enhances the privacy and anonymity of transactions by making it more difficult to trace transactions on the blockchain. 
\item Ponzi:  financial scheme promising high returns to investors by using funds from new investors to pay returns to earlier investors. 
\item Ransomware: malicious software that encrypts files on a victim's computer, demanding payment to decrypt and restore access. 
\item Bridge: protocol that facilitates the exchange of assets between Bitcoin and different blockchain networks (e.g. Ethereum).
\end{itemize}
}

These entities were selected due to their relevance and prevalence within the cryptocurrency ecosystem, providing a comprehensive overview of the diverse actors within the Bitcoin ecosystem. \\

\begin{figure}
    \centering
    \includegraphics[width=0.9\linewidth]{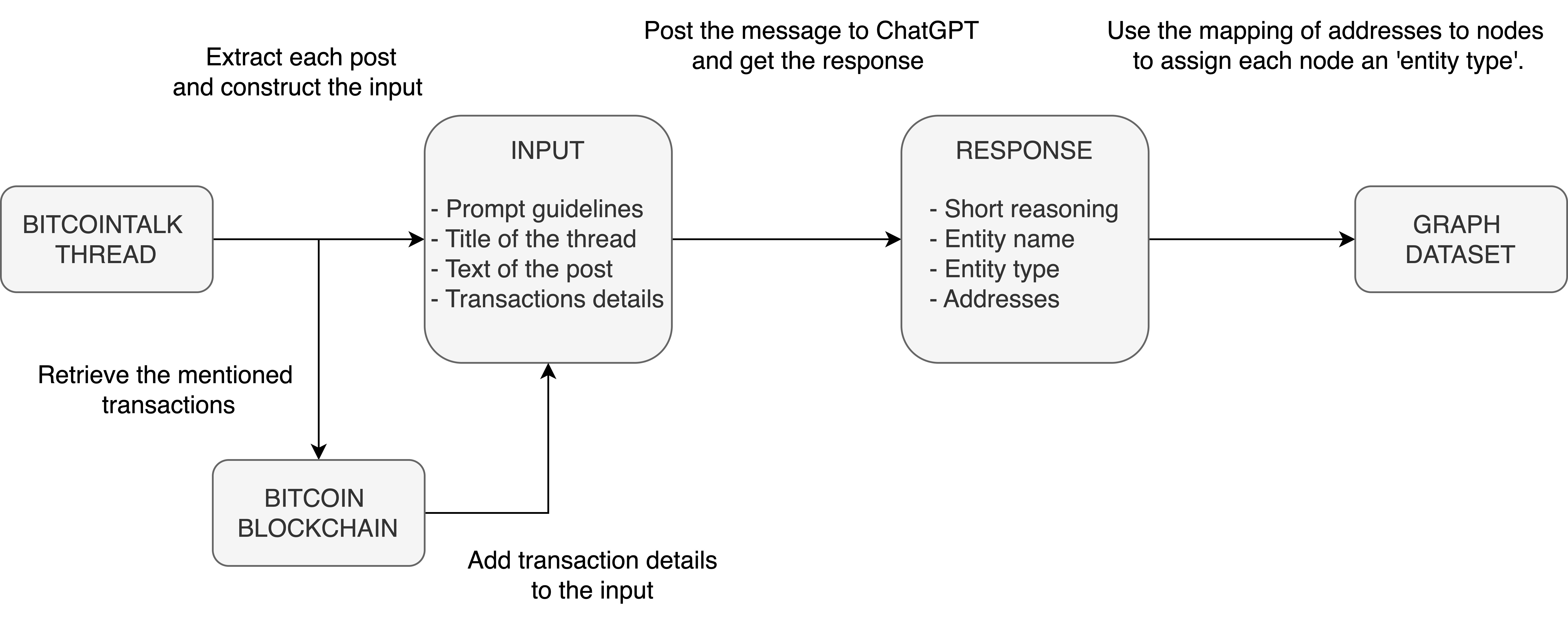}
    \caption{Labeling pipeline.}
    \label{fig:labeling_pipeline}
\end{figure}

The full labeling pipeline is illustrated in Figure \ref{fig:labeling_pipeline}.

\subsubsection*{BitcoinTalk} BitcoinTalk is an online forum dedicated to Bitcoin and remains one of the most active forums on the subject. The forum is divided into several sections, subsections, and threads. A thread is a sequence of messages or posts, which should be related to the thread's topic or title. Bitcoin addresses are often mentioned in posts, and the context of the thread can sometimes help assign these addresses to entities, such as services or companies. To construct the dataset, we extracted messages from the English-speaking part of the forum. We collected the data at the end of 2023. In total, we collected 14,067,713 messages from 546,440 threads. The data collected from BitcoinTalk is included in the dataset (see 'Data Records').

\subsubsection*{ChatGPT}
Addresses will be assigned to entity names using ChatGPT\cite{brown2020language,openai_chatgpt}, an artificial intelligence assistant developed by OpenAI based on the GPT foundation models. ChatGPT is designed to engage in human-like automated conversations with users. The GPT models have been fine-tuned using supervised learning and reinforcement learning from human feedback. A conversation consists of a sequence of user prompts and assistant responses. ChatGPT has demonstrated impressive results in various tasks, including following instructions and solving logic problems. We utilized ChatGPT (model 'gpt-4o-mini') via API calls for this purpose.

\subsubsection*{Deposit Addresses, Hot and Cold Wallets}

We concentrated on addresses owned by organizations, especially those providing services in exchange for bitcoins. Transactions between user addresses and service addresses are relatively common across various services\cite{mccorry2018preventing}.  To access these services, users deposit funds by transferring bitcoins from their personal addresses to addresses managed by the organization, known as a deposit addresses. Organizations typically generate unique deposit addresses for each client, simplifying the monitoring of client deposits. However, these addresses remain under the organization's control. Once funds are deposited from personal addresses to the deposit addresses, users can utilize the service in exchange for the Bitcoins they have deposited. Users can withdraw their remaining Bitcoins to personal addresses after they have finished using the service. Typically, funds from deposit addresses are consolidated into two types of addresses: hot wallets and cold wallets. Hot wallets are internet-connected and hold sufficient funds for routine operations like user withdrawals. Conversely, cold wallets are generally offline to protect against online threats and store the bulk of the service's funds and user deposits. Based on this typical interaction pattern and observations from the collected posts, we can design instructions for ChatGPT to extract information about the addresses mentioned in the posts.

\subsubsection*{Prompts}
We have formulated several prompts to guide ChatGPT in associating addresses mentioned in a post to an entity name, provided that the context (the post and the thread title) allows for it. In addition to the textual information within the post, we incorporate supplementary data. Certain posts include transaction IDs, which serve as unique identifiers of transactions on the blockchain. This enables readers to retrieve detailed transaction information from the blockchain, thereby providing additional context. Since this detailed transaction information can also help ChatGPT, we include transaction details (senders, recipients, amounts) of the mentioned transaction IDs in the prompt. Although all amounts on the blockchain are denominated in satoshis, posters frequently refer to amounts in USD in their posts. To assist ChatGPT in matching amounts in bitcoins in transactions with the USD amounts mentioned in the posts, we added the converted USD amounts using the BTC/USD exchange rate from the date of the post. We used the conversion rate available on CoinMarketCap (\url{https://coinmarketcap.com}). \\

All prompts are available in the code directory (see 'Code availability'). The first script is designed to detect Bitcoin deposits from customers to a service. Posters usually mention deposit addresses or transaction IDs when they encounter issues during their deposit, such as their account not being credited with the correct amount of bitcoins. This script associates mentioned deposit addresses with entity names if the context allows. The second script targets users' withdrawals. When users want to withdraw their funds, they provide a recipient address to the service. The service then creates the transaction and communicates the transaction ID once it is incorporated into the blockchain. Withdrawal transactions are typically funded by the service's funds, likely controlled by hot wallets. For this reason, we assume the sending addresses of these transactions are owned by the identified service. The third script is quite similar; it also attempts to detect withdrawals and aims to identify the involved entity, the address used by the user, and the amount withdrawn. We refer to the previous case by searching the blockchain for the corresponding withdrawal transaction, around the date of the post (+/- three days), with the withdrawal address as the output receiving the indicated amount. If a unique transaction matches these characteristics, we assume it is the withdrawal transaction, and the addresses funding this transaction belong to the detected entity. Finally, the last script identifies hot and cold addresses under various circumstances.

\subsubsection*{Labeling}

Prompts have been meticulously crafted to ensure the assistant returns a succinct reasoning along with an entity name and bitcoin addresses or transaction ids. A mapping between the returned entity names and the predefined entity categories has been established. This mapping process utilized threads from BitcoinTalk where the entity names were mentioned, as well as various internet resources and the Internet Wayback Machine. For each labeled address, we identified and labeled the locking scripts that can be unlocked by the address with the same label than the address. Subsequently, for each labeled script, we assigned the corresponding cluster/node the same label. In instances where a cluster contains multiple conflicting labels, no label was assigned.

\subsubsection*{Limits}
Posts retrieved from BitcoinTalk may contain inaccuracies, misinformation, or deliberate falsehoods posted by users. Additionally, since we only fetched posts from the English-speaking part of the forum, the constructed dataset of labels may not be representative of all entities globally. Moreover,  it is important to highlight the limitations of ChatGPT in inferring entities from short unstructured context. 

\subsubsection*{Other sources}

To enrich our dataset, we incorporated additional data from diverse sources.

\begin{itemize}
    \item We included the addresses of centralized exchanges, labeled as 'exchange', sourced from CoinMarketCap\cite{coinmarketcap_cexs} and DefiLlama\cite{defillama_cexs}. These addresses were provided directly by the exchanges.
    \item  Our collection of ransomware addresses was expanded by incorporating addresses identified in previous research papers, including the Padua ransomware dataset\cite{conti2018economic}, the Sextorsion Padua dataset\cite{ankit2019true}, and the Montréal ransomware dataset\cite{paquet2019ransomware}. These addresses were compiled from public sources such as security reports, academic publications, and online databases listing Bitcoin addresses linked to illegal activities.
    \item We also integrated addresses from the Specially Designated Nationals (SDN) List maintained by the U.S. Department of Justice\cite{sdn_list}. The selected entities from this list include 'Suex', 'Chatex', and 'Garantex' (all labeled as exchanges), 'Hydra' (marketplace), and 'Blender.io' (mixer).
    \item Additionally, we included addresses holding bitcoins related to the bridge Wrapped Bitcoin (WBTC), categorizing them under 'bridge'\cite{wbtc}.
    \item Bitcoin addresses were also extracted from user profiles on BitcoinTalk. Forum users often include their personal Bitcoin addresses in their profiles or signatures, displayed below their posts. We scraped the profiles of all forum posters from previously collected messages, labeling each identified address as 'individual'.
    \item Furthermore, we incorporated addresses associated with mining entities. Miners earn rewards for incorporating new transactions into the blockchain through special transactions known as \textit{Coinbase} transactions. Miners have the ability to include a message in these transactions; some mining companies may embed their name or a distinctive pattern. These messages allow us to identify and categorize these addresses under 'mining'.
    \item Lastly, we included addresses related to betting and gambling platforms. Some platforms, such as Fairlay and DirectBet, enable customers to participate in bets by sending bitcoins to specific addresses created for each bet. Users can share their bets on forums like BitcoinTalk via URLs that redirect to the bet, which include the Bitcoin deposit addresses. We developed regex patterns to detect these URLs, extract the addresses, and classify them as 'bet'.
\end{itemize}

All addresses obtained in this subsection are also included in the dataset (see 'Data Records').

\section*{Data Records}

The latest release of the dataset is accessible at \url{https://figshare.com/articles/dataset/BitcoinTemporalGraph/26305093}, with the DOI \url{https://doi.org/10.6084/m9.figshare.26305093.v1}, under a Creative Commons Attribution 4.0 International license.

\subsection*{BitcoinTalk threads} 

All threads collected from BitcoinTalk and used for labeling are stored in a compressed folder named 'BitcoinTalkThreads.tar.gz'. Each thread is saved as a separate JSON file within this folder. The structure of these JSON files is outlined below. 

\begin{lstlisting}[language=json, xleftmargin=0.1\textwidth, xrightmargin=0.1\textwidth]
{
  "title": "Title of the thread",
  "thread_number": "Unique identifier of the thread on BitcoinTalk",
  "thread_url": "URL of the thread on BitcoinTalk",
  "number_messages": "Number of messages in the thread",
  1: {
    "poster": "Username of the poster of the first message",
    "poster_id": "Unique identifier of the user on BitcoinTalk",
    "date": "Date of the first message", 
    "message": "Content of the first message"
  },
  2: {
    "poster": "Username of the poster of the second message",
    "poster_id": "Unique identifier of the user on BitcoinTalk",
    "date": "Date of the second message", 
    "message": "Content of the second message"
  },
  ...
  }
}
\end{lstlisting}

\begin{figure}
    \centering
    \includegraphics[width=\textwidth]{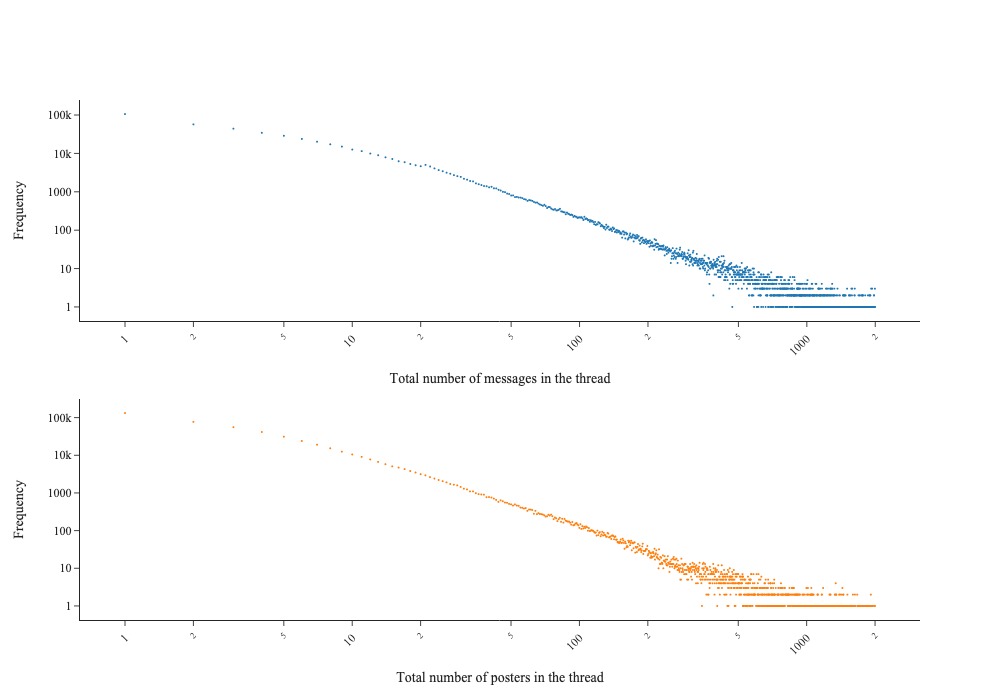}
    \caption{Top: Frequency distribution of the number of messages per thread. Bottom: Frequency distribution of the number of posters per thread.}
    \label{fig:stats_bitcointalk}
\end{figure}

The dataset contains a total of 14,067,713 messages collected from 546,440 threads. On average, each thread contains 26 messages and has contributions from 13 unique posters. Additionally, each message averages 58 words. We have plotted in Figure \ref{fig:stats_bitcointalk} the frequency distributions of the number of messages and posters per thread.

\subsection*{Bitcoin addresses dataset} The file 'addresses.csv' contains all 101,186 labeled addresses obtained during the 'Node labeling' section. In Tables \ref{tab:categories} and \ref{tab:label_sources}, we present the distribution of categories among the labeled addresses, as well as the distribution of label sources.

\begin{table}[h!]
\centering
\begin{minipage}{0.45\textwidth}
\centering
\begin{tabular}{|l|r|r|}
\hline
\textbf{Category}    & \textbf{Count} & \textbf{Percentage} \\ \hline
Individual         & 63,189         & 61\%            \\ \hline
Bet                 & 16,366         & 15.8\%            \\ \hline
Ransomware           & 12,539         & 12.1\%            \\ \hline
Gambling             & 4,718          & 4.55\%             \\ \hline
Exchange             & 2,776          & 2.68\%             \\ \hline
Mining               & 1,608          & 1.55\%             \\ \hline
Ponzi               & 1,012          & 0.97\%             \\ \hline
Marketplace          & 406            & 0.39\%             \\ \hline
Faucet               & 401            & 0.39\%             \\ \hline
Bridge               & 370            & 0.36\%             \\ \hline
Mixer                & 234            & 0.23\%             \\ \hline
\end{tabular}
\caption{Distribution of categories among the labeled addresses.}
\label{tab:categories}
\end{minipage}%
\hfill
\begin{minipage}{0.45\textwidth}
\centering
\begin{tabular}{|l|r|r|}
\hline
\textbf{Label Source}  & \textbf{Count} & \textbf{Percentage} \\ \hline
BitcoinTalk            & 89,246         & 86.1\%            \\ \hline
Montréal               & 7,222          & 6.97\%             \\ \hline
Padua                  & 4,027          & 3.89\%             \\ \hline
PaduaSextorsion        & 1,290          & 1.24\%             \\ \hline
CoinbaseMessages       & 898            & 0.87\%             \\ \hline
WBTC                   & 370            & 0.36\%             \\ \hline
DefiLlama              & 292            & 0.28\%             \\ \hline
SDN                    & 186            & 0.18\%             \\ \hline
CoinMarketCap          & 88             & 0.08\%             \\ \hline
\end{tabular}
\caption{Distribution of label sources.}
\label{tab:label_sources}
\end{minipage}
\end{table}

\subsection*{Graph dataset}

\begin{figure}
    \centering
    \begin{minipage}{0.49\textwidth}
        \centering
        \includegraphics[width=\textwidth]{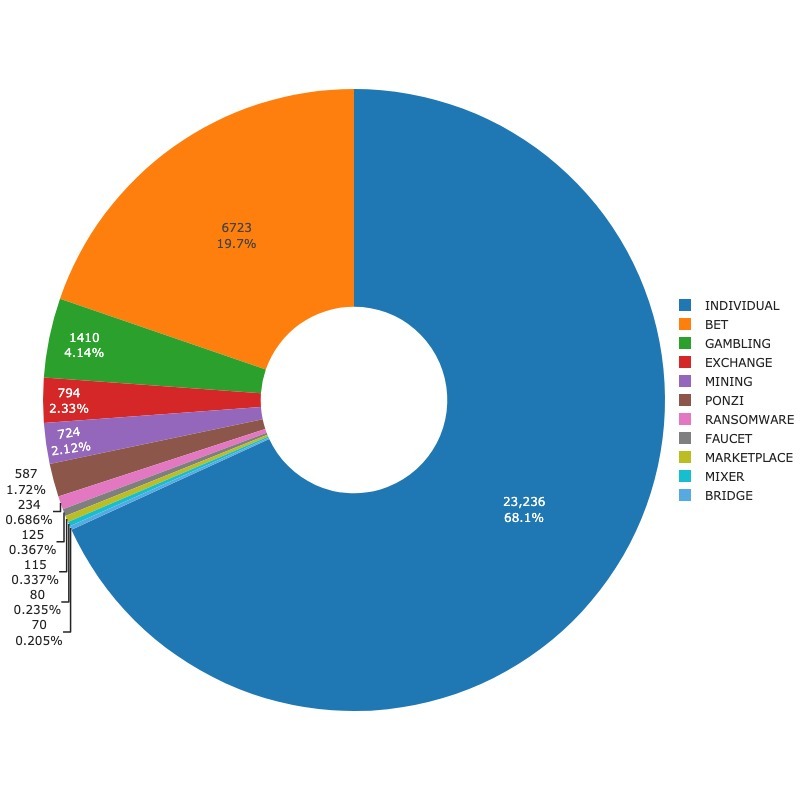}
        \caption{Distribution of categories among the labeled nodes.}
        \label{fig:distribution_category_node}
    \end{minipage}
    \hfill
    \begin{minipage}{0.49\textwidth}
        \centering
        \includegraphics[width=\textwidth]{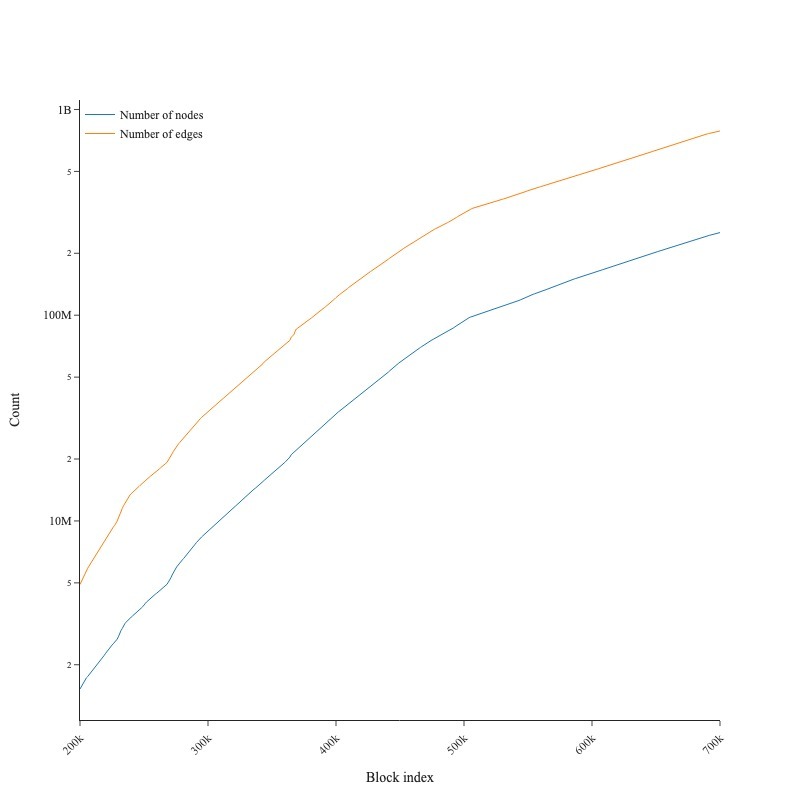} 
        \caption{Evolution of the number of nodes and edges in the graph as a function of the block index.}
        \label{fig:evolution_graph}
    \end{minipage}
\end{figure}

The features of the 252,219,007 nodes and the 785,954,737 edges are stored in a compressed PostgreSQL database, organized in the form of tables. These tables are structured to facilitate efficient querying and analysis of Bitcoin transaction data. The descriptions of these tables are provided in Table \ref{tab:table_edge} and Table \ref{tab:table_node}. The graph's density is approximately 1\%, indicating that it is very sparsely connected. The density of a graph is a measure of the proportion of edges present compared to the maximum possible number of edges in the graph. This suggests that only a small fraction of possible directed edges are present, reflecting a network with minimal direct interactions between nodes. In total, 34,098 nodes are labeled. We have plotted the distribution of categories among the labeled nodes in Figure \ref{fig:distribution_category_node}. We have also plotted in Figure \ref{fig:evolution_graph} the evolution of the number of nodes and edges in the graph as a function of the timestamp (block index in the blockchain).

\subsection*{Other files} The compressed folder 'LabelingOtherSources.tar.gz' contains all the files related to the 'Other sources' subsection of the 'Node labeling' section. This folder includes text files that list various addresses collected from the different alternative sources.

\section*{Technical Validation} 

Predicting the label of a node based on its attributes serves as a validation method for our dataset. This approach establishes a robust connection between off-chain data, the entity type behind a cluster, and on-chain activity data. The successful prediction of these labels underscores the dataset's reliability and relevance, justifying its utility across various applications.

\subsubsection*{Task and Models} The learning task consists of predicting the 'label' column for labeled nodes. We specifically predict the following labels: 'exchange', 'mining', 'gambling', 'ponzi', 'individual', 'ransomware', and 'bet', as these are sufficiently represented in the dataset. For this task, we consider the graph as undirected. We trained four different state-of-the-art graph neural networks (GNNs): GCN\cite{kipf2016semi}, GraphSage\cite{hamilton2017inductive}, GAT\cite{velivckovic2017graph}, and GIN\cite{xu2018powerful}. GNNs can handle complex relationships and dependencies between entities represented as nodes and edges in a graph. They leverage the connectivity patterns and node features to learn representations that capture the structure and features of the graph, making them suitable for tasks like node classification. For these models, we used the implementation provided by the PyTorch Geometric package with the following hyperparameters: 2 layers, ReLU activations, 256 hidden units, and a dropout rate of 0.1. The GAT network was used with 8 attention heads, and we used a 2-layer perceptron with 256 hidden cells to map the node features for the GIN network. To compare the performance of these models with more traditional machine learning approaches, we also trained a gradient boosting classifier (GBC). A GBC is a machine learning algorithm that builds an ensemble of decision trees sequentially to improve prediction accuracy, used on tabular data without relationships between samples, i.e., the feature vectors of the labeled nodes. For this, we used the implementation from the Sklearn package with default parameters. \\

\subsubsection*{Additional features} 
We added features derived from the existing features that appeared useful for our task. We calculated the average amounts received and sent, the proportion of non-isolated addresses within the cluster, the out-degree/in-degree ratio, and the time before the first transaction ($\text{first\_transaction\_out} - \text{first\_transaction\_in}$), if applicable. We also defined the age of a node, as shown in Equation (\ref{eq:age}), as the difference in the number of blocks between its last activity and its first activity.



\begin{equation}
    \text{age} = \text{max}(\text{last\_transaction\_in}, \text{last\_transaction\_out}) - \text{min}(\text{first\_transaction\_in}, \text{first\_transaction\_out})
    \label{eq:age}
\end{equation}

The older an entity, the more likely it is to have created numerous addresses and conducted many transactions. We believe that temporal activity rates of a node are more informative than its absolute activity values. For this reason, we divided certain features by the age of the node, specifically the attributes total\_transaction\_in and total\_transaction\_out, as well as degree and cluster-type attributes.
Even though transaction amounts are in bitcoins, the network users are part of traditional economies. For this reason, Bitcoin users often think in terms of dollars rather than bitcoins, especially given the significant increase in Bitcoin's price since its inception, making it challenging to compare two quantities in Bitcoin. For these reasons, we converted each value-type attribute into USD terms. To achieve this, we used the historical BTC/USD conversion rates to calculate a median satoshi price over the node's activity period.

\subsubsection*{Pre-processing} 
The features exhibited a probability density that followed a power law distribution, except at very large values. To enhance learning and accelerate convergence, we have processed the features using the following pipeline: (1) replacing zero values with NaNs, (2) applying logarithmic transformation, (3) normalization, (4) clipping the values between 0 and 1, and (5) replacing any missing values with 0. For normalization, each feature was adjusted using its minimum value $q_{0\%}$ and the 95th percentile $q_{95\%}$ through the following transformation: $  x \longrightarrow \frac{\log(x) - \log(q_{0\%})}{\log(q_{95\%}) - \log(q_{0\%})} $.
For value-type features, the smaller values did not adhere well to a power law distribution. Therefore, we substituted the minimum value in the aforementioned formula with the 5th percentile. All normalization constants were computed based on the training set.

\subsubsection*{Training} 
We divided the labeled nodes into training, validation, and test splits with distributions of 40\%, 30\%, and 30\%, respectively, and maintained these distributions across all experiments. To address class imbalance and enhance training performance, we employed oversampling for underrepresented labels and undersampling of overrepresented labels, ensuring that each label had between 300 and 1500 samples. Our models were trained using the Adam optimizer to minimize the weighted cross-entropy loss function provided by the Pytorch package. The weights for each label were set inversely proportional to the number of samples of that label in the training set, with the weights normalized so that the average is 1. We set the batch size to 32 and the learning rate to $10^{-4}$, and $3 \times 10^{-4}$ for the GAT network. \\

\subsubsection*{Neighborhood sampling} 
The constructed graph is large-scale, making it challenging to train graph neural network models. Additionally, the number of labeled nodes is relatively small compared to the graph's size ($\approx$0.014\%). For each labeled node, we used only sampled neighborhood around each node, resulting in a dataset consisting of subgraphs sampled from the initial graph around each labeled node. The sampling is characterized by a maximum depth $k_\text{max}$ and a number of neighbors to explore at each depth. The process is as follows: given a depth $1 \leq k \leq k_\text{max}$, a number of neighbors to be sampled $n_k$, a group of already explored nodes $\mathcal{N}_\text{prev}$, and a group of nodes to explore, $\mathcal{N}_\text{next}$. For each node $n$ to explore, denote $\mathcal{V}(n)$ as the neighbors of $n$ that have neither been explored nor are currently being explored. If the number of neighbors is less than $n_k$, all these neighbors are selected for the next depth; otherwise, $n_k$ neighbors are selected without replacement. At the end of this loop, the nodes explored at the current depth are considered fully explored ($\mathcal{N}_\text{prev} \leftarrow \mathcal{N}_\text{prev} \cup \mathcal{N}_\text{next}$). The group of nodes to explore in the next depth, i.e., $\mathcal{N}_\text{next}$, consists of all new sampled neighbors. The neighborhood of a labeled node is obtained by starting the exploration from the labeled node $n_\text{seed}$. The neighborhood of a labeled node is thus the graph composed of all explored nodes and the nodes to be explored in the last depth, i.e., $\mathcal{N}_\text{next} \cup \mathcal{N}_\text{prev}$. We add each edge between a node and its parent in the exploration. The sampling process is illustrated in Algorithm \ref{alg:neighborhood_sampling}. In our experiments, we used a maximum depth of $k_\text{max} = 2$, and the number of neighbors to sample were 10 and 5 for the first and second steps, respectively.  \\

\begin{algorithm}[H]
\caption{Neighborhood sampling pseudo-code}\label{alg:neighborhood_sampling}
\KwIn{$n_\text{seed}$: a labeled node, $k_\text{max} \geq 1$: the maximal depth, $n_1, n_2, \dots n_{k_\text{max}}$: the number of neighbors to explore at each depth.}
\KwOut{$\mathcal{G}$: a neighborhood of $n_\text{seed}$.}

$\mathcal{E} \gets \{\}$\; \tcp*{Edges of the output graph}
$\mathcal{N}_\text{prev} \gets \{\}$\; \tcp*{Already explored}
$\mathcal{N}_\text{next} \gets \{n_\text{seed}\}$\; \tcp*{To be explored}

\For{$k \gets 1$ \KwTo $k_\text{max}$}{
    $\hat{\mathcal{N}}_\text{next} \gets \{\}$\; \tcp*{To be explored in the next loop}
    
    \For{$n \in \mathcal{N}_\text{next}$}{
    
        $\mathcal{V}(n)$: the neighbors of $n$\; 
        
        $\mathcal{V}(n) \gets \mathcal{V}(n) - (\mathcal{N}_\text{prev} \cup \mathcal{N}_\text{next} \cup \hat{\mathcal{N}}_\text{next})$\;

        \If{$|\mathcal{V}(n)| > n_k$}{
            $\hat{\mathcal{V}}(n)$: a sample from $\mathcal{V}(n)$ without replacement of size $n_k$\; 
        }
        \Else{
            $\hat{\mathcal{V}}(n) \gets {\mathcal{V}}(n)$\;
        }
        $\hat{\mathcal{N}}_\text{next} \gets \hat{\mathcal{N}}_\text{next} \cup \hat{\mathcal{V}}(n)$\;  \tcp*{Add the selected neighbors}
        
        \For{$n^\prime \in \hat{\mathcal{N}}(n)$}{  
            $\mathcal{E} \gets \mathcal{E} \cup \{(n, n^\prime), (n^\prime, n) \} $\;  \tcp*{Add the new edges to the output graph}
        }
    
}
$\mathcal{N}_\text{prev} \gets \mathcal{N}_\text{prev} \cup \mathcal{N}_\text{next}$\; 

$\mathcal{N}_\text{next} \gets \hat{\mathcal{N}}_\text{next}$\; 
}

$\mathcal{N} \gets \mathcal{N}_\text{prev} \cup \mathcal{N}_\text{next}$\; \tcp*{Nodes of the output graph}

$\mathcal{G} \gets (\mathcal{N}, \mathcal{E})$\; \tcp*{Output graph}

\Return{$\mathcal{G}$}
\end{algorithm}

Finding the neighbors to explore for a node involves extracting all edges connected to the node from the dataset and deducing its neighborhood. This operation is particularly resource-intensive and time-consuming for high-degree nodes. To speed up the process, for nodes with a degree greater than 100,000, we requested only a random sample of the edges connected to the node. To achieve this, we used the sampling method 'TABLESAMPLE SYSTEM' in PostgreSQL. \\

\subsubsection*{Data augmentation}

The neighborhood sampling process is inherently stochastic. By sampling multiple neighborhoods from the same labeled node, we effectively increase the total number of samples available for training and evaluation. This method of data augmentation can enhance the model's generalization capacity, and provides a more comprehensive assessment of the model's performance and robustness. Given the computational expense of the sampling process relative to model training, we have implemented a buffer of pre-sampled neighborhoods. For each split of the labeled nodes (train, validation, and test), we created 12 split-datasets. Each split-dataset consists of a sampled neighborhood for each node belonging to the split. The normalization constants for preprocessing are derived from a randomly selected training split-dataset prior to training. During the training phase, train and validation split-datasets were periodically (every 100 epochs) randomly selected from the buffer to continue the training process.

\subsubsection*{Results analysis} For each node in the test split, and for each neighborhood of this node, stored in the buffer, we computed the probability vector for each class using the trained GNNs. Subsequently, we calculated the mean probability vector by averaging the probability vectors obtained from all neighborhoods of the node. The final classification label for each node was determined by selecting the class with the highest mean probability. We compared the performance of different models using F1 score for each label, and the average F1 score (macro-F1) over the different classes. The results presented in Table \ref{tab:performance_comparison}. The F1 score is a measure of predictive performance for classification models, accounting for both false positives and false negatives, especially useful in imbalanced datasets. It ranges from 0 to 1, where 1 indicates perfect classification, and 0 indicates the worst performance. We also included the performances of the Gradient Boosting Classifier (GBC). \\ 

Based on the presented results, node features are useful for predicting the entity type of a node. For instance, the Gradient Boosting Classifier achieves a macro-F1 score of 0.57, with relatively high F1 scores for the 'gambling', 'individual', and 'bet' classes. The higher scores of GraphSage, GAT, and GIN suggest that a node's neighborhood is useful for prediction, which is particularly true for the 'mining', 'ponzi', and 'bet' classes. The GAT and GIN models, known for their high expressiveness, achieve the best results, with macro-F1 scores of 0.64 and 0.63, respectively. In contrast to the Gradient Boosting Classifier, GNNs struggle significantly to predict the 'ransomware' class. \\

\begin{table}[]
\footnotesize
\centering
\begin{tabular}{|l|c|ccccccc|}
\hline
\textbf{Model}  & \textbf{Macro-F1} & \textbf{F1 Exchange} & \textbf{F1 Mining} & \textbf{F1 Gambling} & \textbf{F1 Ponzi} & \textbf{F1 Individual} & \textbf{F1 Ransomware} & \textbf{F1 Bet} \\
\hline
GCN & 0.56 & 0.43 & 0.45 & 0.56 & 0.46 & 0.88 & 0.19 & \textbf{0.98}   \\ \hline
GraphSage & 0.62 & 0.51 & \textbf{0.57} & 0.58 & 0.55 & 0.93 & 0.25 & \textbf{0.98}  \\ \hline
GAT & \textbf{0.64} & 0.49 & \textbf{0.57} & \textbf{0.62} & 0.57 & 0.93 & 0.32 & \textbf{0.98}  \\ \hline
GIN & 0.63 & \textbf{0.54} & 0.54 & \textbf{0.62} & \textbf{0.60} & \textbf{0.94} & 0.21 & \textbf{0.98}  \\ \hline
GBC  & 0.57 & \textbf{0.54} & 0.45 & 0.57 & 0.30 & \textbf{0.94} & \textbf{0.50} & 0.89 \\
\hline
\end{tabular}
\caption{Performance comparison of the trained models on test split.}
\label{tab:performance_comparison}
\end{table}

We have plotted in Table \ref{tab:confusion_matrix} the normalized confusion matrix for the trained GAT model. Each entry in the matrix is calculated by dividing the number of instances of the true label (row) predicted as the column label by the total number of instances of the true label (row). There is a noticeable pattern of misclassification from various labels to 'individual', suggesting that the model may need further refinement in distinguishing 'individual' from other categories.

\begin{table}[]
    \centering
    \footnotesize
    \begin{tabular}{|c|ccccccc|}
        \hline
         & \textbf{Exchange} & \textbf{Mining} & \textbf{Gambling} & \textbf{Ponzi} & \textbf{Individual} & \textbf{Ransomware} & \textbf{Bet} \\ \hline
        \textbf{Exchange} & 64\% & 3\% & 7\% & 4\% & 19\% & 2\% & 1\% \\ \hline
        \textbf{Mining} & 5\% & 70\% & 3\% & 3\% & 17\% & 1\% & 0\% \\ \hline
        \textbf{Gambling} & 7\% & 1\% & 68\% & 6\% & 15\% & 1\% & 3\% \\ \hline
        \textbf{Ponzi} & 5\% & 1\% & 4\% & 75\% & 14\% & 0\% & 1\% \\ \hline
        \textbf{Individual} & 3\% & 2\% & 3\% & 2\% & 88\% & 2\% & 0\% \\ \hline
        \textbf{Ransomware} & 14\% & 0\% & 3\% & 0\% & 23\% & 60\% & 0\% \\ \hline
        \textbf{Bet} & 0\% & 0\% & 1\% & 0\% & 0\% & 0\% & 99\% \\ \hline
    \end{tabular}
    \caption{Row-normalized confusion matrix of the GAT model. True labels are represented by the rows, and predicted labels are represented by the columns. }
    \label{tab:confusion_matrix}
\end{table}



\section*{Usage Notes}

\subsection*{Restoring the graph database} 

After decompressing the archive, the tables can be restored into an existing PostgreSQL database using the \textit{pg\_restore} utility as follows: \verb|pg_restore -j number_jobs -Fd -0 -U database_username -d database_name dataset|. In this command, \verb|number_jobs| is the number of threads to use to load the data and create indexes, \verb|database_username| is the is the username to connect as, and \verb|database_name| is the name of the PostgreSQL database. Once the data is loaded into the database, indexes on certain columns will be automatically created to accelerate data access in downstream tasks. The databse requires substantial storage: 40GB for node\_features and 80GB for transaction\_edges (including indexes). We recommend provisioning sufficient storage to accommodate these tables. In our research, we configured several PostgreSQL parameters to optimize database performance. Specifically, we set shared\_buffers to 1,048,576 (default: 16,384), work\_mem to 16,384 (default: 4,096), maintenance\_work\_mem to 1,048,576 (default: 65,536), wal\_buffers to 2,048 (default: -1), and max\_parallel\_workers\_per\_gather to 4 (default: 2). We recommend matching or exceeding these configurations to achieve optimal performance.

\subsection*{Use cases}
We have demonstrated  how to use the dataset to perform entity type prediction. There are some other exciting research directions as yet to be explored by using this dataset.

\begin{itemize}
    \item \textit{Study of interactions among different entity types based on value flows.} The temporal dimension of the graph can be instrumental in determining the extent to which these interactions are modulated by political, regulatory, economic or financial contexts. Previous research\cite{huang2018tracking,paquet2019spams,gomez2022watch} has for instance particularly examined the value flows involving ransomwares and  other cyber criminal activites, elucidating how extorted funds are subsequently laundered. 
    \item \textit{Evolution of the graph over time.} This encompasses an analysis of the changes in the number of nodes and edges\cite{filtz2017evolution,alqassem2018anti}, the connectivity / density of the graph, as well as the identification of dynamic zones over time. The graph delineates interactions among users from the inception of the network, capturing various phases of adoption of this emergent economy—from its marginal utilization to the near-institutionalization of Bitcoin as a financial asset.
    \item \textit{Comparison with other economies and networks.} Prior studies\cite{alqassem2018anti,tao2021complex} have delved into the topological characteristics of the interaction graph among Bitcoin users. These characteristics can subsequently be compared to those of other well-known networks, thereby enhancing the understanding of Bitcoin. In addition, the large scale of this graph is advantageous for the pre-training of neural networks, which can be subsequently fine-tuned for other graphs involving transactions, payments, or more broadly, value transfers between real entities.
\end{itemize}

\section*{Code availability}

The code used in this research is available on GitHub. The repositories are detailed as follows:
\begin{itemize}
    \item \textbf{BTCGraphConstruction}: This repository contains the code for constructing the PostgreSQL database where the graph tables are stored. It can be accessed at \url{https://github.com/hugoschnoering2/BTCGraphConstruction}.
    \item \textbf{BTCGraphLabeling}: This repository includes the code for labeling Bitcoin addresses based on data collected from Bitcoin Talk threads and other data sources. The repository is available at \url{https://github.com/hugoschnoering2/BTCGraphLabeling}.
    \item \textbf{BTCGraphPredictingLabel}: This repository hosts the code for the prediction task presented in the Technical Validation section of the paper. It can be found at \url{https://github.com/hugoschnoering2/BTCGraphPredictingLabel}.
\end{itemize}

These repositories contain all the necessary code and documentation to replicate the results and methodologies discussed in this paper.

\bibliography{sample}

\section*{Acknowledgements} 

Hugo Schnoering acknowledges the financial support of Coinshares for funding his PhD.

\section*{Author contributions} 

The first author conducted the engineering work needed (collection, cleaning, exploration of the data, experimental evaluation) and wrote the paper, the second author supervised the work contributed to the design of the experimental evaluation and contributed to reviewing the paper.

\section*{Competing interests} 

The authors declare no competing interests.

\end{document}

%% file: transaction_scheme.tex
\begin{figure}[h]

\centering

\resizebox{0.99\textwidth}{!}{
\begin{tikzpicture}

\node[circle, draw, font=\tiny, minimum size=0.50cm] (A) at (-7.1, 3.5) {};
\node[font=\tiny] (A) at (-6.3, 3.5) {TXO};

\node[circle, draw, font=\tiny, minimum size=0.45cm] (A) at (-5.5, 3.5) {};
\node[circle, draw, font=\tiny, minimum size=0.50cm] (A) at (-5.5, 3.5) {};
\node[font=\tiny] (A) at (-4.4, 3.5) {Transaction};

\node[circle, draw, font=\tiny, minimum size=0.50cm] (TXOA) at (-3.2, 3.5) {A};
\node[circle, draw, font=\tiny, minimum size=0.50cm] (transationB) at (-2.2, 3.5) {B};
\node[circle, draw, font=\tiny, minimum size=0.45cm] (A) at (-2.2, 3.5) {};
\draw[arrows = {-Stealth[length=6pt,fill=none]}] (TXOA) -- (transationB);

\node[font=\tiny] (A) at (-0., 3.5) {[TXO A] funds [Transaction B]};

\node[circle, draw, font=\tiny, minimum size=0.50cm] (transationC) at (2.2, 3.5) {C};
\node[circle, draw, font=\tiny, minimum size=0.45cm] (A) at (2.2, 3.5) {};
\node[circle, draw, font=\tiny, minimum size=0.50cm] (TXOD) at (3.2, 3.5) {D};
\draw[arrows = {-Stealth[length=6pt]}] (transationC) -- (TXOD);

\node[font=\tiny] (A) at (5.5, 3.5) {[Transaction C] creates [TXO D]};

\draw [decorate, decoration = {brace}] (-5.5,-2) --  (-5.5,2);
\draw [decorate, decoration = {brace}] (5,2) --  (5,-2);
\node[circle, font=\tiny] (prevtransaction) at (-7,0) {Previous transactions};
\node[circle, font=\tiny] (prevtransaction) at (6.5,0) {Future transactions};


\node[circle, draw, font=\tiny, minimum size=0.72cm] (transactionin1) at (-5,2) {};
\node[circle, draw, font=\tiny, minimum size=0.65cm] (transactionin1bis) at (-5,2) {};
\node[circle, draw, font=\tiny, minimum size=0.72cm] (transactionin2) at (-5,1) {};
\node[circle, draw, font=\tiny, minimum size=0.65cm] (transactionin2bis) at (-5,1) {};
\node[circle, draw, font=\tiny, minimum size=0.72cm] (transactionink) at (-5,-2) {};
\node[circle, draw, font=\tiny, minimum size=0.65cm] (transactioninibis) at (-5,-2) {};

\node[circle, draw, font=\tiny] (nodein1) at (-2,2) {$p_1$};
\node[circle, draw, font=\tiny] (nodein2) at (-2,1) {$p_2$};
\node[circle, font=\tiny] (nodeinj) at (-2,-0.5) {$\vdots$};
\node[circle, draw, font=\tiny] (nodeink) at (-2,-2) {$p_k$};

\node[circle, draw, font=\Large, minimum size=0.95cm] (nodedelta) at (0,0) {};
\node[circle, draw, font=\Large, minimum size=0.8cm] (nodedelta1) at (0,0) {$\Delta$};

\node[circle, font=\tiny] (transactionout1) at (5,2) {};
\node[circle, font=\tiny] (transactionout2) at (5,1) {};
\node[circle, font=\tiny] (transactionoutk) at (5,-2) {};

\node[circle, draw, font=\tiny] (nodeout1) at (2,2) {$p_1^\prime$};
\node[circle, draw, font=\tiny] (nodeout2) at (2,1) {$p_2^\prime$};
\node[circle, font=\tiny] (nodeoutj) at (2,-0.5) {$\vdots$};
\node[circle, draw, font=\tiny] (nodeoutk) at (2,-2) {$p_l^\prime$};


\draw[arrows = {-Stealth[length=8pt]}, dotted] (transactionin1) -- (nodein1) node[midway, above,font=\tiny] {$v_1$};
\draw[arrows = {-Stealth[length=8pt]}, dotted] (transactionin2) -- (nodein2) node[midway, above,font=\tiny] {$v_2$};
\draw[arrows = {-Stealth[length=8pt]}, dotted] (transactionink) -- (nodeink) node[midway, above,font=\tiny] {$v_k$};

\draw[arrows = {-Stealth[fill=none,length=8pt]}] (nodein1) -- (nodedelta) node[midway, above,font=\tiny] {$v_1$};
\draw[arrows = {-Stealth[fill=none,length=8pt]}] (nodein2) -- (nodedelta) node[midway, above,font=\tiny] {$v_2$};
\draw[arrows = {-Stealth[fill=none,length=8pt]}] (nodeink) -- (nodedelta) node[midway, above,font=\tiny] {$v_k$};

\draw[arrows = {-Stealth[fill=none,length=8pt]}, dotted] (nodeout1) -- (transactionout1) node[midway, above,font=\tiny] {$v_1^\prime$};
\draw[arrows = {-Stealth[fill=none,length=8pt]}, dotted] (nodeout2) -- (transactionout2) node[midway, above,font=\tiny] {$v_2^\prime$};
\draw[arrows = {-Stealth[fill=none,length=8pt]}, dotted] (nodeoutk) -- (transactionoutk) node[midway, above,font=\tiny] {$v_k^\prime$};

\draw[arrows = {-Stealth[length=8pt]}] (nodedelta) -- (nodeout1) node[midway, above,font=\tiny] {$v_1^\prime$};
\draw[arrows = {-Stealth[length=8pt]}] (nodedelta) -- (nodeout2) node[midway, above,font=\tiny] {$v_2^\prime$};
\draw[arrows = {-Stealth[length=8pt]}] (nodedelta) -- (nodeoutk) node[midway, above,font=\tiny] {$v_l^\prime$};

\end{tikzpicture}
}
\caption{Schematic of a transaction $\Delta$. Nodes with a single (resp. double) border represent TXOs (resp. transactions). TXOs consumed by $\Delta$ originate from prior transactions, while those created in $\Delta$ may serve as input TXOs in subsequent transactions.}
\label{fig:scheme_transaction}

\end{figure}

%% file: main.bbl
\begin{thebibliography}{10}
\urlstyle{rm}
\expandafter\ifx\csname url\endcsname\relax
  \def\url#1{\texttt{#1}}\fi
\expandafter\ifx\csname urlprefix\endcsname\relax\def\urlprefix{URL }\fi
\expandafter\ifx\csname doiprefix\endcsname\relax\def\doiprefix{DOI: }\fi
\providecommand{\bibinfo}[2]{#2}
\providecommand{\eprint}[2][]{\url{#2}}

\bibitem{nakamoto2008bitcoin}
\bibinfo{author}{Nakamoto, S.}
\newblock \bibinfo{journal}{\bibinfo{title}{Bitcoin: A peer-to-peer electronic cash system}}.
\newblock {\emph{\JournalTitle{Satoshi Nakamoto}}}  (\bibinfo{year}{2008}).

\bibitem{glassnode}
\bibinfo{author}{{Glassnode}}.
\newblock \bibinfo{title}{Glassnode studio bitcoin}.
\newblock \bibinfo{howpublished}{\url{https://studio.glassnode.com/metrics?a=BTC}} (\bibinfo{year}{2024}).
\newblock \bibinfo{note}{Accessed: 2024-05-01}.

\bibitem{googlescholar}
\bibinfo{author}{{Google}}.
\newblock \bibinfo{title}{Google scholar}.
\newblock \bibinfo{howpublished}{\url{https://scholar.google.com/scholar?q=bitcoin&as_ylo=2023&as_yhi=2023}} (\bibinfo{year}{2023}).
\newblock \bibinfo{note}{Accessed: 2024-05-01}.

\bibitem{conti2018economic}
\bibinfo{author}{Conti, M.}, \bibinfo{author}{Gangwal, A.} \& \bibinfo{author}{Ruj, S.}
\newblock \bibinfo{journal}{\bibinfo{title}{On the economic significance of ransomware campaigns: A bitcoin transactions perspective}}.
\newblock {\emph{\JournalTitle{Computers \& Security}}} \textbf{\bibinfo{volume}{79}}, \bibinfo{pages}{162--189} (\bibinfo{year}{2018}).

\bibitem{paquet2019ransomware}
\bibinfo{author}{Paquet-Clouston, M.}, \bibinfo{author}{Haslhofer, B.} \& \bibinfo{author}{Dupont, B.}
\newblock \bibinfo{journal}{\bibinfo{title}{Ransomware payments in the bitcoin ecosystem}}.
\newblock {\emph{\JournalTitle{Journal of Cybersecurity}}} \textbf{\bibinfo{volume}{5}}, \bibinfo{pages}{tyz003} (\bibinfo{year}{2019}).

\bibitem{weber2019anti}
\bibinfo{author}{Weber, M.} \emph{et~al.}
\newblock \bibinfo{journal}{\bibinfo{title}{Anti-money laundering in bitcoin: Experimenting with graph convolutional networks for financial forensics}}.
\newblock {\emph{\JournalTitle{arXiv preprint arXiv:1908.02591}}}  (\bibinfo{year}{2019}).

\bibitem{bellei2024shape}
\bibinfo{author}{Bellei, C.} \emph{et~al.}
\newblock \bibinfo{journal}{\bibinfo{title}{The shape of money laundering: Subgraph representation learning on the blockchain with the elliptic2 dataset}}.
\newblock {\emph{\JournalTitle{arXiv preprint arXiv:2404.19109}}}  (\bibinfo{year}{2024}).

\bibitem{schnoering2024assessing}
\bibinfo{author}{Schnoering, H.}, \bibinfo{author}{Porthaux, P.} \& \bibinfo{author}{Vazirgiannis, M.}
\newblock \bibinfo{journal}{\bibinfo{title}{Assessing the efficacy of heuristic-based address clustering for bitcoin}}.
\newblock {\emph{\JournalTitle{arXiv preprint arXiv:2403.00523}}}  (\bibinfo{year}{2024}).

\bibitem{schnoering2023heuristics}
\bibinfo{author}{Schnoering, H.} \& \bibinfo{author}{Vazirgiannis, M.}
\newblock \bibinfo{journal}{\bibinfo{title}{Heuristics for detecting coinjoin transactions on the bitcoin blockchain}}.
\newblock {\emph{\JournalTitle{arXiv preprint arXiv:2311.12491}}}  (\bibinfo{year}{2023}).

\bibitem{colored_coins}
\bibinfo{author}{{Bitcoin Wiki}}.
\newblock \bibinfo{title}{Colored coins}.
\newblock \bibinfo{howpublished}{\url{https://en.bitcoin.it/wiki/Colored_Coins}} (\bibinfo{year}{2024}).
\newblock \bibinfo{note}{Accessed: 2024-07-19}.

\bibitem{brown2020language}
\bibinfo{author}{Brown, T.} \emph{et~al.}
\newblock \bibinfo{journal}{\bibinfo{title}{Language models are few-shot learners}}.
\newblock {\emph{\JournalTitle{Advances in neural information processing systems}}} \textbf{\bibinfo{volume}{33}}, \bibinfo{pages}{1877--1901} (\bibinfo{year}{2020}).

\bibitem{openai_chatgpt}
\bibinfo{author}{{OpenAI}}.
\newblock \bibinfo{title}{Introducing chatgpt}.
\newblock \bibinfo{howpublished}{\url{https://openai.com/index/chatgpt/}} (\bibinfo{year}{2024}).
\newblock \bibinfo{note}{Accessed: 2024-07-19}.

\bibitem{mccorry2018preventing}
\bibinfo{author}{McCorry, P.}, \bibinfo{author}{M{\"o}ser, M.} \& \bibinfo{author}{Ali, S.~T.}
\newblock \bibinfo{title}{Why preventing a cryptocurrency exchange heist isn’t good enough}.
\newblock In \emph{\bibinfo{booktitle}{Security Protocols XXVI: 26th International Workshop, Cambridge, UK, March 19--21, 2018, Revised Selected Papers 26}}, \bibinfo{pages}{225--233} (\bibinfo{organization}{Springer}, \bibinfo{year}{2018}).

\bibitem{coinmarketcap_cexs}
\bibinfo{author}{{CoinMarketCap}}.
\newblock \bibinfo{title}{Coinmarketcap cexs}.
\newblock \bibinfo{howpublished}{\url{https://coinmarketcap.com/rankings/exchanges/}} (\bibinfo{year}{2024}).
\newblock \bibinfo{note}{Accessed: 2024-02-02}.

\bibitem{defillama_cexs}
\bibinfo{author}{{DefiLlama}}.
\newblock \bibinfo{title}{Defillama cexs}.
\newblock \bibinfo{howpublished}{\url{https://defillama.com/cexs}} (\bibinfo{year}{2024}).
\newblock \bibinfo{note}{Accessed: 2024-02-04}.

\bibitem{ankit2019true}
\bibinfo{author}{Gangwal, A.}
\newblock \bibinfo{title}{The true scale of extortions by sextortion emails}, \url{10.13140/RG.2.2.33765.24808} (\bibinfo{year}{2019}).

\bibitem{sdn_list}
\bibinfo{author}{{US Department of Justice}}.
\newblock \bibinfo{title}{Sdn list}.
\newblock \bibinfo{howpublished}{\url{https://sanctionslist.ofac.treas.gov/Home/SdnList}} (\bibinfo{year}{2023}).
\newblock \bibinfo{note}{Accessed: 2023-12-20}.

\bibitem{wbtc}
\bibinfo{author}{{Wrapped Bitcoin (WBTC)}}.
\newblock \bibinfo{title}{Wbtc audit}.
\newblock \bibinfo{howpublished}{\url{https://wbtc.network/dashboard/audit}} (\bibinfo{year}{2023}).
\newblock \bibinfo{note}{Accessed: 2023-12-20}.

\bibitem{kipf2016semi}
\bibinfo{author}{Kipf, T.~N.} \& \bibinfo{author}{Welling, M.}
\newblock \bibinfo{journal}{\bibinfo{title}{Semi-supervised classification with graph convolutional networks}}.
\newblock {\emph{\JournalTitle{arXiv preprint arXiv:1609.02907}}}  (\bibinfo{year}{2016}).

\bibitem{hamilton2017inductive}
\bibinfo{author}{Hamilton, W.}, \bibinfo{author}{Ying, Z.} \& \bibinfo{author}{Leskovec, J.}
\newblock \bibinfo{journal}{\bibinfo{title}{Inductive representation learning on large graphs}}.
\newblock {\emph{\JournalTitle{Advances in neural information processing systems}}} \textbf{\bibinfo{volume}{30}} (\bibinfo{year}{2017}).

\bibitem{velivckovic2017graph}
\bibinfo{author}{Veli{\v{c}}kovi{\'c}, P.} \emph{et~al.}
\newblock \bibinfo{journal}{\bibinfo{title}{Graph attention networks}}.
\newblock {\emph{\JournalTitle{arXiv preprint arXiv:1710.10903}}}  (\bibinfo{year}{2017}).

\bibitem{xu2018powerful}
\bibinfo{author}{Xu, K.}, \bibinfo{author}{Hu, W.}, \bibinfo{author}{Leskovec, J.} \& \bibinfo{author}{Jegelka, S.}
\newblock \bibinfo{journal}{\bibinfo{title}{How powerful are graph neural networks?}}
\newblock {\emph{\JournalTitle{arXiv preprint arXiv:1810.00826}}}  (\bibinfo{year}{2018}).

\bibitem{huang2018tracking}
\bibinfo{author}{Huang, D.~Y.} \emph{et~al.}
\newblock \bibinfo{title}{Tracking ransomware end-to-end}.
\newblock In \emph{\bibinfo{booktitle}{2018 IEEE Symposium on Security and Privacy (SP)}}, \bibinfo{pages}{618--631} (\bibinfo{organization}{IEEE}, \bibinfo{year}{2018}).

\bibitem{paquet2019spams}
\bibinfo{author}{Paquet-Clouston, M.}, \bibinfo{author}{Romiti, M.}, \bibinfo{author}{Haslhofer, B.} \& \bibinfo{author}{Charvat, T.}
\newblock \bibinfo{title}{Spams meet cryptocurrencies: Sextortion in the bitcoin ecosystem}.
\newblock In \emph{\bibinfo{booktitle}{Proceedings of the 1st ACM conference on advances in financial technologies}}, \bibinfo{pages}{76--88} (\bibinfo{year}{2019}).

\bibitem{gomez2022watch}
\bibinfo{author}{Gomez, G.}, \bibinfo{author}{Moreno-Sanchez, P.} \& \bibinfo{author}{Caballero, J.}
\newblock \bibinfo{title}{Watch your back: identifying cybercrime financial relationships in bitcoin through back-and-forth exploration}.
\newblock In \emph{\bibinfo{booktitle}{Proceedings of the 2022 ACM SIGSAC conference on computer and communications security}}, \bibinfo{pages}{1291--1305} (\bibinfo{year}{2022}).

\bibitem{filtz2017evolution}
\bibinfo{author}{Filtz, E.}, \bibinfo{author}{Polleres, A.}, \bibinfo{author}{Karl, R.} \& \bibinfo{author}{Haslhofer, B.}
\newblock \bibinfo{title}{Evolution of the bitcoin address graph: An exploratory longitudinal study}.
\newblock In \emph{\bibinfo{booktitle}{Data Science--Analytics and Applications: Proceedings of the 1st International Data Science Conference--iDSC2017}}, \bibinfo{pages}{77--82} (\bibinfo{organization}{Springer}, \bibinfo{year}{2017}).

\bibitem{alqassem2018anti}
\bibinfo{author}{Alqassem, I.}, \bibinfo{author}{Rahwan, I.} \& \bibinfo{author}{Svetinovic, D.}
\newblock \bibinfo{journal}{\bibinfo{title}{The anti-social system properties: Bitcoin network data analysis}}.
\newblock {\emph{\JournalTitle{IEEE Transactions on Systems, Man, and Cybernetics: Systems}}} \textbf{\bibinfo{volume}{50}}, \bibinfo{pages}{21--31} (\bibinfo{year}{2018}).

\bibitem{tao2021complex}
\bibinfo{author}{Tao, B.} \emph{et~al.}
\newblock \bibinfo{journal}{\bibinfo{title}{Complex network analysis of the bitcoin transaction network}}.
\newblock {\emph{\JournalTitle{IEEE Transactions on Circuits and Systems II: Express Briefs}}} \textbf{\bibinfo{volume}{69}}, \bibinfo{pages}{1009--1013} (\bibinfo{year}{2021}).

\end{thebibliography}
